# HazardNet: A Small-Scale Vision Language Model for Real-Time Traffic Safety Detection at Edge Devices

Mohammad Abu Tami
*Natural, Engineering and Technology Sciences Department*
*Arab American University*
Jenin, Palestine
mabutame@gmail.com

Mohammed Elhenawy
*Centre for Accident Research and Road Safety (CARRS-Q),*
*Queensland University of Technology*
Kelvin Grove, Australia.
mohammed.elhenawy@qut.edu.au

Huthaifa I. Ashqar
*AI and Data Science Department*
*Arab American University*
Jenin, Palestine
huthaifa.ashqar@aaup.edu

*Abstract*—Traffic safety remains a vital concern in contemporary urban settings, intensified by the increase of vehicles and the complicated nature of road networks. Traditional safety-critical event detection systems predominantly rely on sensor-based approaches and conventional machine learning algorithms, necessitating extensive data collection and complex training processes to adhere to traffic safety regulations. This paper introduces HazardNet, a small-scale Vision Language Model designed to enhance traffic safety by leveraging the reasoning capabilities of advanced language and vision models. We built HazardNet by fine-tuning the pre-trained Qwen2-VL-2B model, chosen for its superior performance among open-source alternatives and its compact size of two billion parameters. This helps to facilitate deployment on edge devices with efficient inference throughput. In addition, we present HazardQA, a novel Vision Question Answering (VQA) dataset constructed specifically for training HazardNet on real-world scenarios involving safety-critical events. Our experimental results show that the fine-tuned HazardNet outperformed the base model up to an 89% improvement in F1-Score and has comparable results with improvement in some cases reach up to 6% when compared to larger models, such as GPT-4o. These advancements underscore the potential of HazardNet in providing real-time, reliable traffic safety event detection, thereby contributing to reduced accidents and improved traffic management in urban environments. Both HazardNet model and the HazardQA dataset are available at https://huggingface.co/Tami3/HazardNet and https://huggingface.co/datasets/Tami3/HazardQA, respectively.

## I. INTRODUCTION

Traffic safety is a critical issue in rapidly urbanizing areas, with road traffic accidents causing approximately 1.35 million deaths annually worldwide [1]. Effective traffic safety management not only saves lives but also reduces economic losses associated with accidents, such as healthcare costs, property damage, and loss of productivity [2]. Traditional safety-critical event detection systems rely on sensor-based approaches and conventional machine learning (ML) algorithms, which demand extensive annotated data and struggle to generalize across the diverse and dynamic conditions of urban traffic [3,4]. These systems often operate in isolation, limiting their effectiveness in comprehensive traffic management.

Recent advancements in Large Language Models (LLMs) and Vision Language Models (VLMs) introduce promising enhancements for traffic safety systems via integrating multiple data modalities such as text and images, which improve event detection and decision-making [5,6]. However, the large size and computational demands of these models hold their deployment on edge devices necessary for real-time applications [7].

This paper introduces HazardNet, a small-scale Vision Language Model, which fine-tuned from the pre-trained Qwen2-VL-2B model [8], selected for its superior performance and compact size of 2 billion parameters, enabling efficient deployment on edge devices. Additionally, we present HazardQA, a novel Vision Question Answering dataset designed to train HazardNet on real-world safety-critical scenarios. Experimental results show that HazardNet outperforms base model with enhancement reached up to 89% and has comparable results with improvement in some cases that reached up to 6% in larger models like GPT-4o [9]. The remainder of this paper is structured as follows: Section 2 reviews related literature, Section 3 details the methods used, Section 4 presents the experimental studies and results, and Section 5 concludes with future work.

## II. LITERATURE REVIEW

The integration of LLMs into autonomous driving systems has shown significant potential in enhancing decision-making, perception, and interaction [10]. LLM4Drive study [11] highlight how LLMs improve these areas through Chain-of-Thought (CoT) reasoning and contextual understanding, categorizing research into planning, perception, question answering, and generation while tackling challenges like transparency and scalability. Similarly, Cui et al. [12] examine the combination of LLMs with vision foundation models (VFMs), tracing the evolution from sensor-based to deep learning techniques that enhance perception and decision-making, and reviewing essential datasets such as KITTI [13] and nuScenes [14]. Driving with llms [15] introduced a pretraining method aligning numeric vectors with LLM representations, enhancing scenario interpretation and decision-making.

Advanced frameworks like DriveMLM [16] and "Drive As You Speak" [17] demonstrate the alignment of multimodal LLMs with behavioral planning and natural language interactions, respectively, while AccidentGPT [18] uses multimodal models for comprehensive traffic accident analysis. A recent study [19] have focused on in-context learning (ICT) for automated detection of traffic safety-critical events. Others [20, 21] integrate real-time sensor data with LLMs to boost autonomous driving (AD) functionalities and improve object detection and pedestrian behavior prediction by combining LLMs with LiDAR, radar, and contextual information. On another hand, existing datasets such as KITTI [13], Cityscapes [22], and the Waymo Open Dataset [23] have substantially advanced research in traffic safety and autonomous driving by providing large-scale, high-resolution images and sensor data with diverse driving conditions. Likewise, the DRAMA dataset [24] offers real-world footage focusing on driver attention and anomalies, further emphasizing the importance of robust perception in complex road environments.

Although AD systems have advanced, they still struggle with corner cases [25], hampering zero-shot performance. Existing data-driven methods [26-28] and multimodal large language models [29-31] fail to provide adequate generalization, leading to the need for a custom VLM optimized for real-time detection on edge devices. Moreover, current datasets support tasks such as object detection and segmentation but lack question-based annotations essential for deeper understanding and interaction regarding traffic safety events. This limitation underscores the necessity of an easy-to-use VQA dataset specifically designed for critical, real-world traffic scenarios, enabling enhanced semantic comprehension and interactive reasoning for safer autonomous driving systems.

This body of work highlights two main issues that our study seeks to address. First, there is an evident gap in existing datasets, which lack the question-based annotations necessary for achieving a nuanced semantic understanding of traffic safety events. Second, while significant advances have been made, current methods do not adequately generalize across the dynamic and diverse conditions encountered in urban settings, nor do they facilitate efficient edge deployment. These limitations motivate our development of HazardNet—a compact and efficient Vision Language Model optimized for edge devices—and the introduction of HazardQA, a novel Vision Question Answering dataset specifically designed to capture real-world safety-critical scenarios.

## III. METHODOLOGY

This section outlines the methodologies employed in developing HazardNet, focusing on the creation of the HazardQA dataset and the fine-tuning process of the base model using Low-Rank Adaptation (LoRA) [32] and Quantized LoRA (QLoRA) [33].

### A. Dataset Creation

The HazardQA dataset was developed to train HazardNet on real-world scenarios involving safety-critical events. This dataset is derived from the existing human-annotated DRAMA dataset, which focuses on Joint Risk Localization and Captioning in Driving [24]. DRAMA contains ~17K distinct real-world scenarios as illustrated in Fig. 1. Each sample contains the frame taken from the ego-car with annotated data that describe the scene such as: caption, is there a risk, suggested action and road classification among others.

We extend each scenario in DRAMA by generating multiple question-answer (QA) pairs, resulting in a new dataset of safety-critical QA interactions as explained in Algorithm 1. The core idea in creating this dataset is to:

- Extract the relevant annotations from DRAMA for each scenario.
- Prompt an LLM, specifically GPT-4o, with these annotations.
- Generate five question-answer pairs (QA pairs) per scenario (as shown in Fig. 2), ensuring that each pair captures a distinct safety or risk-related aspect of the driving context.

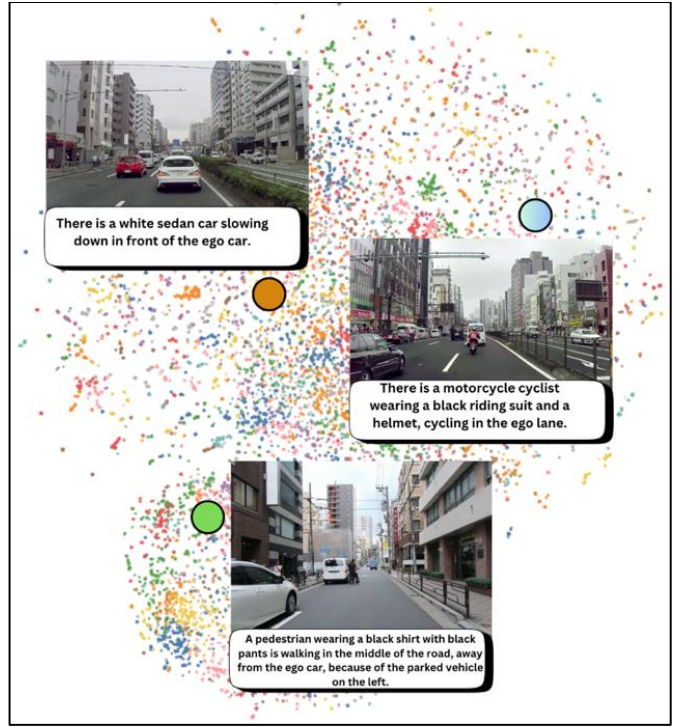

Fig. 1. DRAMA dataset Embeddings projected into 2D-Map can be found at https://atlas.nomic.ai/data/mabutame/drama-image-embedding2/map/.

Algorithm 1: Creating HazardQA.
**Input:** DRAMA dataset with N annotated scenarios.
**Output:** HazardQA dataset with 5N VQA pairs.
1. **Initialize** HazardQA ← { }
2. **For i = 1 to** N **do**
    a. Retrieve the annotated data for scenario **i** from DRAMA:(caption, risk presence, suggested action, road type, etc).
    b. Construct a prompt for **GPT-4o** that includes:
        i. The scenario description using the DRAMA annotations.
        ii. Instructions to generate 5 diverse QA pairs.
    c. Use GPT-4o to obtain the set of 5 QA pairs {(Q1,A1), ..., (Q5, A5)}.
    d. Append the resulting 5 QA pairs to HazardQA.
3. **End** for
4. **Return** HazardQA.

To contextualize our newly created HazardQA dataset, TABLE I provides a comparative overview of our newly introduced HazardQA dataset alongside several well-known autonomous driving and traffic safety datasets, highlighting differences in scale, annotation type, task focus, and safety-critical aspects.

*B. Model Training*

To learn from the newly created HazardQA dataset, we train a model by fine-tuning a vison language model via parameter-efficient strategies. Specifically, we employ LoRA and QLoRA to reduce computational overhead and memory usage while still achieving high performance.

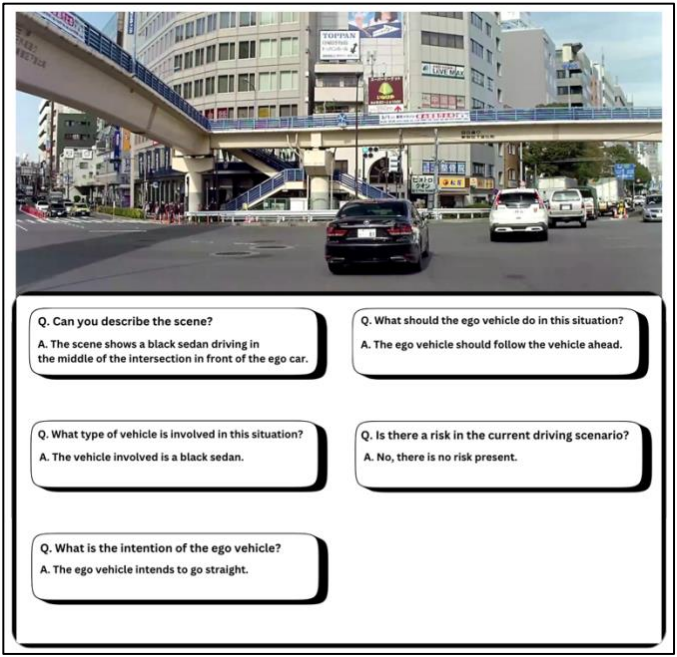

Fig. 2. An Example Scene from HazardQA Showcasing Five Safety-Critical QA Pairs.

LoRA injects trainable, low-rank modifications into a pre-trained model's weight matrices without altering the original weights $W$. Formally, let $W \in \mathbb{R}^{d_{in} \times d_{out}}$ be a frozen weight matrix in the Transformer (e.g., within the self-attention or feed-forward layers). LoRA parameterizes the update $\Delta W$ using two low-rank matrices $A$ and $B$ can be shown as:

$$\Delta W = AB \qquad (1)$$

where $B \in \mathbb{R}^{d_{in} \times r}$ and $A \in \mathbb{R}^{r \times d_{out}}$, with $r \ll \min(d_{in}, d_{out})$. Consequently, the updated weight matrix becomes:

$$W' = W + \Delta W = W + BA \qquad (2)$$

During training, only the low-rank parameters ($A$, $B$) are updated, while $W$ remains fixed as illustrated in Fig. 3. This low-rank adaptation drastically decreases the number of learnable parameters, making fine-tuning significantly more efficient.

QLoRA extends LoRA by applying quantization to the base model weights $W$. Specifically, the large model is quantized into a 4-bit representations as follows:

$$W_{4-bit} = Quant_4(W) \qquad (3)$$

During the forward pass, a dequantization step $DeQuant_4(.)$ reconstructs approximate values of $W$:

$$\overline{W} = DeQuant_4(W_{4-bit}) \qquad (4)$$

Next, the LoRA update $\Delta W = BA$ is added in higher precision (e.g., FP16), yielding:

$$W' = \overline{W} + \Delta W = DeQuant_4(W_{4-bit}) + BA \qquad (5)$$

By freezing the quantized weights $W_{4-bit}$ and training only the low-rank parameters ($A$, $B$), QLoRA combines the memory savings of a 4-bit quantization with the fine-grained adaptability of LoRA, thus enabling efficient fine-tuning on resource-constrained hardware.

TABLE I. COMPARISON OF HAZARDQA WITH EXISTING AUTONOMOUS DRIVING AND TRAFFIC SAFETY DATASETS.

| Comparison | HazardQA | NuScenes [14] | CityScape [22] | Waymo Open [23] | DRAMA [24] |
|---|---|---|---|---|---|
| Task | Visual Question Answering (safety-critical QA) | 3D Detection, Tracking, Planning | Semantic & Instance Segmentation | 3D Detection, Tracking, Prediction | Joint Risk Localization and Captioning |
| Size | ~17K images (each with 5 QA pairs, ~85K QA) | 1k | 5K | ~1k segments (20s each) | ~17K |
| Modality | Monocular RGB images (plus, textual QA pairs) | RGB images, LiDAR, RADAR plus metadata | RGB images | RGB images, LiDAR, multiple cameras, GPS, IMU | RGB images (text annotations) |
| Annotation | - Question-Answer pairs<br>- Safety/risk contexts<br>- Scene-level info | - 3D bounding boxes<br>- Object tracking<br>- Sensor fusion data | - Pixel-level semantic<br>- Segmentation<br>- Instance segmentation | - 3D bounding boxes<br>- Object tracking<br>- Sensor fusion data | - Bounding boxes<br>- Risk indicators<br>- Captions<br>- Scene attributes |
| QA-based? | **Yes** | No | No | No | No |
| Focus on safety Risk? | Yes | Partial | No | Partial | Yes (risk detection, captioning) |

## IV. EXPERIMENTS AND RESULTS

This section presents the experimental setup, evaluation metrics, baseline models, and the results obtained from evaluating HazardNet against exisiting foundational models in traffic safety event detection. The experiments aim to demonstrate the efficacy of HazardNet in accurately classifying safety-critical events while maintaining computational efficiency suitable for real-time deployment.

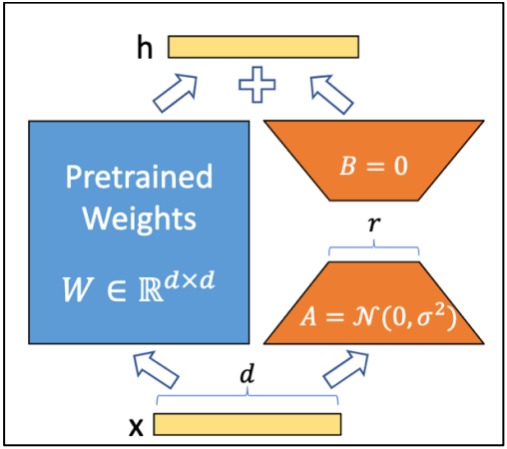

Fig. 3. LoRA reparameterization [32].

Given that the testing task involves multiclass classification, accuracy, recall, precision and f1-score metrics were employed to evaluate the performance of HazardNet and the baseline models. To contextualize the performance of HazardNet, two baseline models were selected for comparison:

1. **GPT4o-mini:** A smaller variant of GPT-4o with over 8 billion parameters, serving as a reference for Multimodal large language models.
2. **Pre-trained Qwen2-VL-2B:** The base model used for fine-tuning HazardNet, consisting of 2 billion parameters.

All models were evaluated on a subset of 500 samples from the testing set, adhering to the classification tasks outlined above.

### A. Training Configuration and Results

Fine-tuning of HazardNet was optimized for both performance and efficiency using the AdamW-8bit optimizer, which reduces memory usage while maintaining effective weight updates. A learning rate of $2e^{-4}$ was selected to ensure stable convergence without overshooting. Low-Rank Adaptation parameters were set with a rank $r = 16$ and a scaling factor $\alpha = 16$ balancing model capacity and computational efficiency. A training batch size of two and gradient accumulation steps of four were employed to optimize GPU memory usage. Additionally, a linear learning rate scheduler with five warmup steps and a weight decay of 0.01 were implemented to enhance generalization and prevent overfitting. The model was trained for one epoch on an NVIDIA A100 GPU, only about 2,179,072 parameters were fine-tuned, representing a mere 0.16% of the base model's parameters, and the entire process was completed in approximately 20 minutes.

The performance of HazardNet was benchmarked against GPT4o-mini and the pre-trained Qwen2-VL-2B model across four classification tasks. Results are summarized in TABLE II.

The analysis of results demonstrates that HazardNet consistently outperforms the pre-trained Qwen2-VL-2B model across most classification tasks, particularly in precision and recall metrics, indicating enhanced accuracy in identifying safety-critical events. While HazardNet slightly beat the larger GPT4o-mini in some measures, it achieves comparable performance in most of the cases while it has significantly fewer parameters, which highlight its efficiency. Notably, HazardNet excels in Agent Classification and Is Risk tasks, showing substantial improvements over baselines. These findings underscore HazardNet's ability to effectively balance performance and efficient computation, making it highly suitable for real-time traffic safety deployment.

TABLE II. EVALUATION OF HAZARDNET AGAINST BASELINE MODELS.

| VQA | Metric | Model | | |
|---|---|---|---|---|
| | | *GPT4o* | *Qwen2-VL-2B* | *HazardNet* |
| **Scene** | Accuracy | 60.00 | 49.52 | 54.29 |
| | Recall | 67.13 | 68.66 | 73.26 |
| | Precision | 60.00 | 49.52 | 54.29 |
| | F1-score | 59.18 | 39.61 | 46.86 |
| **Agent** | Accuracy | 60.95 | 12.38 | 41.9 |
| | Recall | 50.65 | 11.66 | 43.25 |
| | Precision | 60.95 | 12.38 | 41.90 |
| | F1-score | 54.04 | 8.87 | 41.25 |
| **Suggestion Action** | Accuracy | 40.95 | 14.29 | 43.39 |
| | Recall | 28.33 | 8.59 | 30.33 |
| | Precision | 40.95 | 14.29 | 43.37 |
| | F1-score | 28.93 | 8.91 | 35.70 |
| **Risk** | Accuracy | 89.52 | 10.48 | 88.57 |
| | Recall | 86.07 | 1.10 | 80.05 |
| | Precision | 89.52 | 10.48 | 88.57 |
| | F1-score | 86.14 | 1.99 | 84.10 |

## V. CONCLUSION

This paper introduces HazardNet, a small-scale Vision Language Model trained for real-time traffic safety event detection. By fine-tuning the efficient Qwen2-VL-2B model and utilizing the newly developed HazardQA dataset, HazardNet achieves significant performance improvements over existing models while maintaining computational efficiency suitable for edge deployment. Experimental results demonstrate HazardNet's precision and balanced recall across various classification tasks, highlighting its potential to enhance urban traffic management and reduce accident rates. Additionally, by making both the model and dataset publicly available on HuggingFace, this work fosters further research and innovation in traffic safety systems. Future work will explore expanding HazardQA and integrating additional data modalities to further boost HazardNet's capabilities.